\documentclass[conference]{IEEEtran}
\IEEEoverridecommandlockouts

\usepackage{multirow}
\usepackage {subfig}
\usepackage {caption}
\newcommand{\myPara}[1]{\vspace{0.001in}\noindent\textbf{#1}}
\usepackage{url}            
\usepackage{xcolor}         
\usepackage{amsthm,amsmath,amssymb}
\usepackage{array}
\usepackage{mathrsfs}
\usepackage{multirow}
\usepackage{makecell}
\def\ie{{\em i.e.}}
\def\eg{{\em e.g.}}

\usepackage{cite}
\usepackage{amsmath,amssymb,amsfonts}
\usepackage{algorithmic}
\usepackage{graphicx}
\usepackage{textcomp}
\usepackage{xcolor}
\def\BibTeX{{\rm B\kern-.05em{\sc i\kern-.025em b}\kern-.08em
    T\kern-.1667em\lower.7ex\hbox{E}\kern-.125emX}}

\begin{document}

\title{MambaTrack: Exploiting  Dual-Enhancement for Night UAV Tracking }

\author{\IEEEauthorblockN{Chunhui~Zhang$^{1,2,3}$, Li~Liu$^{2,*}$\thanks{*Corresponding author: avrillliu@hkust-gz.edu.cn.}, Hao~Wen$^{3}$, Xi~Zhou$^{1,3}$, Yanfeng~Wang$^{1,4}$}

\IEEEauthorblockA{$^{1}$\textit{Cooperative Medianet Innovation Center, Shanghai Jiao Tong University, Shanghai, China}\\
$^{2}$\textit{Hong Kong University of Science and Technology (Guangzhou), Guangzhou, China}\\
$^{3}$\textit{CloudWalk Technology, Shanghai, China}~~~~~
$^{4}$\textit{Shanghai AI Laboratory, Shanghai, China}
} 
}

\maketitle

\begin{abstract}
Night unmanned aerial vehicle (UAV) tracking is impeded by the challenges of poor illumination, with previous daylight-optimized methods demonstrating suboptimal performance in low-light conditions, limiting the utility of UAV applications. To this end, we propose an efficient mamba-based tracker, leveraging dual enhancement techniques to boost night UAV tracking. The mamba-based low-light enhancer, equipped with an illumination estimator and a damage restorer, achieves global image enhancement while preserving the details and structure of low-light images. Additionally, we advance a cross-modal mamba network to achieve efficient interactive learning between vision and language modalities. Extensive experiments showcase that our method achieves advanced performance and exhibits significantly improved computation and memory efficiency. For instance, our method is 2.8$\times$ faster than CiteTracker and reduces 50.2$\%$ GPU memory. Our codes are available at {\color{darkgray} {\url{https://github.com/983632847/Awesome-Multimodal-Object-Tracking}}}. 


\end{abstract}


\vspace{-0.1cm}
\section{Introduction}
\label{sec:intro}
\vspace{-0.1cm}
Visual tracking, a fundamental task in computer vision, has received widespread attention in various unmanned aerial vehicle (UAV) applications, \eg, aerial photography, target following, smart agriculture, and delivery. It aims to sequentially track a moving object in a video from its initial location. Despite significant advancements, recent efforts have largely concentrated on daytime UAV tracking~\cite{webuav3m2023tpami,li2022all,zhang2020accurate}. Consequently, numerous UAV-related applications are severely hindered due to the insufficient exploration of night UAV tracking~\cite{ye2022unsupervised,zhang2023comprehensive}.

Generally, video frames captured at night have lower brightness and contrast compared to daytime~\cite{li2022all,ge2020cascaded,ge2019distilling,zhang2024segment}. To pursue better tracking performance in the dark night, current works mainly employ two strategies, \ie, low-light enhancement~\cite{ye2021darklighter,ye2022tracker,li2021adtrack} and unsupervised domain adaptation (UDA)~\cite{ye2022unsupervised,zhang2023progressive}. Among them, ADTrack~\cite{li2021adtrack} is a pioneering anti-darkness UAV tracker that adopts a tone mapping algorithm within a correlation filter-based framework to enhance tracking robustness in low-light conditions. Later, some works proposed to eliminate the influence of poor illumination by multiple iterations~\cite{ye2021darklighter} or a spatial-channel Transformer~\cite{ye2022tracker}. However, these low-light enhancement-based methods either struggle to preserve important local details of the image or have a high computational burden. In contrast, UDA-based methods~\cite{ye2022unsupervised,zhang2023progressive} propose to learn from unlabeled data in the target domain (\ie, nighttime scenarios), and utilize knowledge from the source domain (\ie, daytime scenarios) to improve performance in the target domain. However, these methods are challenged by the imbalanced source/target distribution~\cite{zhang2023progressive} and limited data~\cite{zhang2022visible} for night UAV tracking.

\begin{figure}[t]
\centering
\centerline{\includegraphics[width=1.0\linewidth]{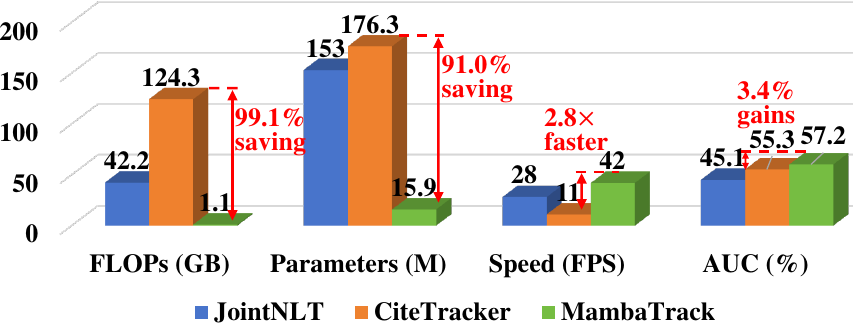}}
\caption{Performance and efficiency comparisons between the proposed MambaTrack and two SOTA trackers (\ie, JointNLT~\cite{zhou2023joint} and CiteTracker~\cite{li2023citetracker}) on UAVDark135~\cite{li2022all}.}
\label{fig:motivation}
\vspace{-0.5cm}
\end{figure}

To alleviate above problems, we propose an efficient mamba-based~\cite{gu2023mamba} tracker (MambaTrack), which includes two simple and effective mechanisms, namely a mamba-based low-light enhancement and a cross-modal enhancement. Owing to the linear computational complexity of the mamba model, our tracker exhibits significant efficiency compared to state-of-the-art (SOTA) methods (see Fig.~\ref{fig:motivation}). Inspired by the recent success of state space models (\eg, Mamba~\cite{gu2023mamba}) in natural language, speech, and various computer vision tasks, we introduce a mamba-based low-light enhancer to improve tracking performance in low-light videos. The low-light enhancer can achieve global enhancement and preserve the local structure of low-light images. Moreover, from a novel multimodal learning view, we annotate language prompts for existing night UAV tracking datasets~\cite{ye2022tracker,ye2022unsupervised,li2021adtrack,li2022all} for semantic enhancement to mitigate the issue of limited nighttime data. This can greatly reduce the cost of annotating massive dense bounding boxes.

\begin{figure*}[t]
\centering
\centerline{\includegraphics[width=1.0\linewidth]{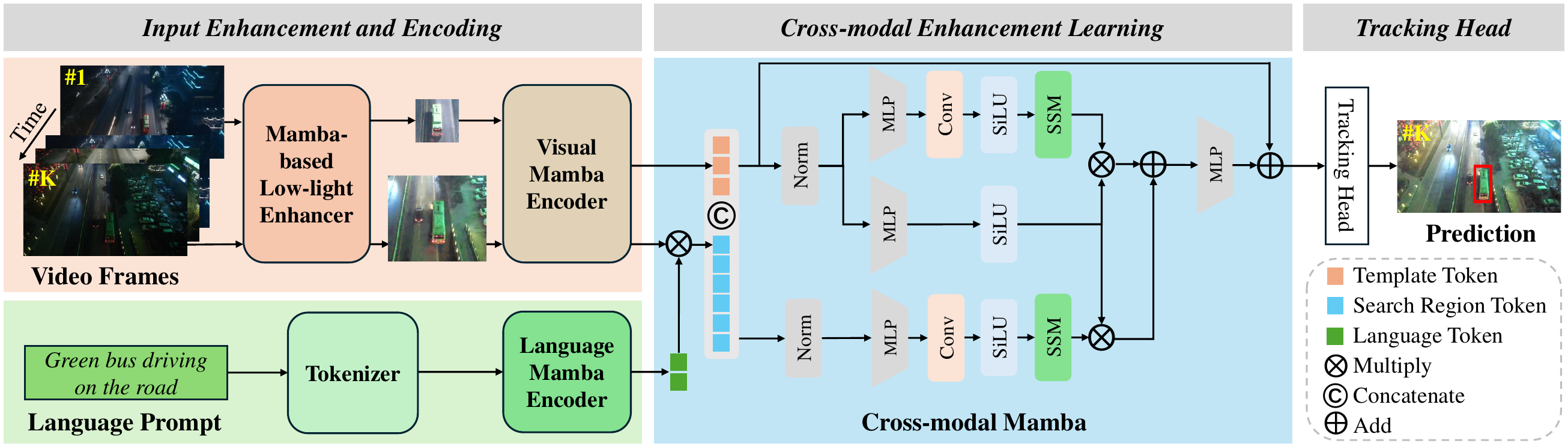}}
\caption{\textbf{Overview of MambaTrack.} It comprises visual and language branches (left), a cross-modal mamba network (middle), and a tracking head (right). The visual branch mainly contains a mamba-based low-light enhancer and a visual mamba encoder for image enhancement and encoding, respectively. The language branch includes a tokenizer and a language mamba encoder. Then, we adopt a cross-modal mamba network for multimodal enhancement learning. Finally, the language-enhanced search embeddings are fed into the tracking head to predict the target. For simplicity, linear projections are omitted here.
}
\label{fig:mambatrack}
\vspace{-0.6cm}
\end{figure*}

In summary, the contributions of this work are fourfold:
\vspace{-0.1cm}
\begin{itemize}
    \item We propose the \textbf{first} mamba-based baseline tracker for night UAV tracking, achieving advanced performance with high computation and memory efficiency.

    \item We present a mamba-based low-light enhancer (MLLE) and a cross-modal mamba (CMM) network for image enhancement and language (semantic) enhancement.

    \item We introduce a \textbf{new} vision-language night UAV tracking task by adding language prompts for existing datasets.

    \item The proposed approach has been evaluated on five challenging benchmarks and has shown its effectiveness.
\end{itemize}

\section{Proposed Approach}
\label{sec:approach}

\subsection{Input Representation}
\label{sec:input_representation}

As shown in Fig.~\ref{fig:mambatrack}, the inputs for MambaTrack contain video frames $\{I_i\}_{i=1}^{N}$, and a language prompt $S$, where $N$ denotes the number of frames. In the visual branch, we crop a small template $Z$ from the initial frame $I_1$ and a search area $X$ from the subsequent frames $\{I_i\}_{i=2}^{N}$. The template and search region can be enhanced using the proposed mamba-based low-light enhancer. Subsequently, a linear projection is employed to obtain 1D tokens $H^{o}_z\in \mathbb{R}^{N_z\times D_1}$ and $H^{o}_x\in \mathbb{R}^{N_x\times D_1}$, with $N_z$ and $N_x$ denoting the length of the tokens, and $D_1$ representing the token dimension. These tokens are then fed into the visual mamba encoder to obtain visual embeddings $H_z$, $H_x$ of the template and search region. In the language branch, the language prompt $S$ is first fed into a text tokenizer to obtain language tokens. Similar to~\cite{zhou2023joint,zhang2023all}, a class token $[\rm CLS]$ is added to the beginning of the language tokens. Subsequently, the language tokens are fed into the language mamba encoder to obtain language embeddings $H_t\in \mathbb{R}^{N_t\times D_2}$, with $N_t$ and $D_2$ denoting the length and dimension of the tokens. Finally, we use a linear projection to transform the dimensions of language embeddings to $D_1$.

\subsection{Mamba-based Low-light Enhancer} 

Based on Retinex theory~\cite{land1977retinex}, a low-light image $I\in \mathbb{R}^{h\times w\times 3}$ can be decomposed into a reflectance image $R\in \mathbb{R}^{h\times w\times 3}$ and an illumination map $L\in \mathbb{R}^{h\times w}$, as $I=R\otimes L$, where $\otimes$ denotes element-wise multiplication. $R$ is determined by the essential properties of the target, and $L$ represents lighting conditions. Following~\cite{cai2023retinexformer}, we introduce perturbations to model noise and artifacts in low-light conditions, as follows:
\begin{equation}
    I=(R+\hat{R})\otimes(L+\hat{L}),
\label{eq:retinex}
\end{equation}
where $\hat{R}\in \mathbb{R}^{h\times w\times 3}$ and $\hat{L}\in \mathbb{R}^{h\times w}$ represent perturbations of $R$ and $L$. Given a light-up map $\bar{L}$, with $\bar{L}\otimes L=1$~\cite{cai2023retinexformer}, we can calculate the light-up image $I_{lu}$ as follows:
\begin{equation}
    I_{lu}=I\otimes \bar{L}=R+R\otimes (\hat{L}\otimes \bar{L})+\hat{R}\otimes(L+\hat{L})\otimes \bar{L},
\label{eq:light-up}
\end{equation}
where $R$ can be considered a well-exposed image and $C=R\otimes (\hat{L}\otimes \bar{L})+\hat{R}\otimes(L+\hat{L})\otimes \bar{L}$ is the total corruption term. Eq.~(\ref{eq:light-up}) can be simplified as:
\begin{equation}
    I_{lu}=I\otimes \bar{L}=R+C.
\label{eq:simplified}
\end{equation}
Therefore, the final enhanced image $I_{en}\in \mathbb{R}^{h\times w\times 3}$ can be calculated as follows:
\begin{equation}
\left\{
\begin{aligned}
(I_{lu}, F_{lu})=IE(I, L_{p}),\\
I_{en}=I_{lu}+DR(I_{lu}, F_{lu}),
\end{aligned}
\right.
\label{eq:mamba-low-light}
\end{equation}
where IE and DR denote the illumination estimator and damage restorer, $L_{p}\in \mathbb{R}^{h\times w}$ represents the illumination prior, $F_{lu}\in \mathbb{R}^{h\times w\times d}$ denotes the light-up feature map, where the channel dimension $d$ for $F_{lu}$ is 40. In our approach, $L_{p}$ is the average value of the channels~\cite{cai2023retinexformer} of the low-light image used to measure the brightness of the image.

\begin{figure*}[ht]
\vspace{-0.2cm}
\centering
\subfloat{\includegraphics[width =0.5\columnwidth]{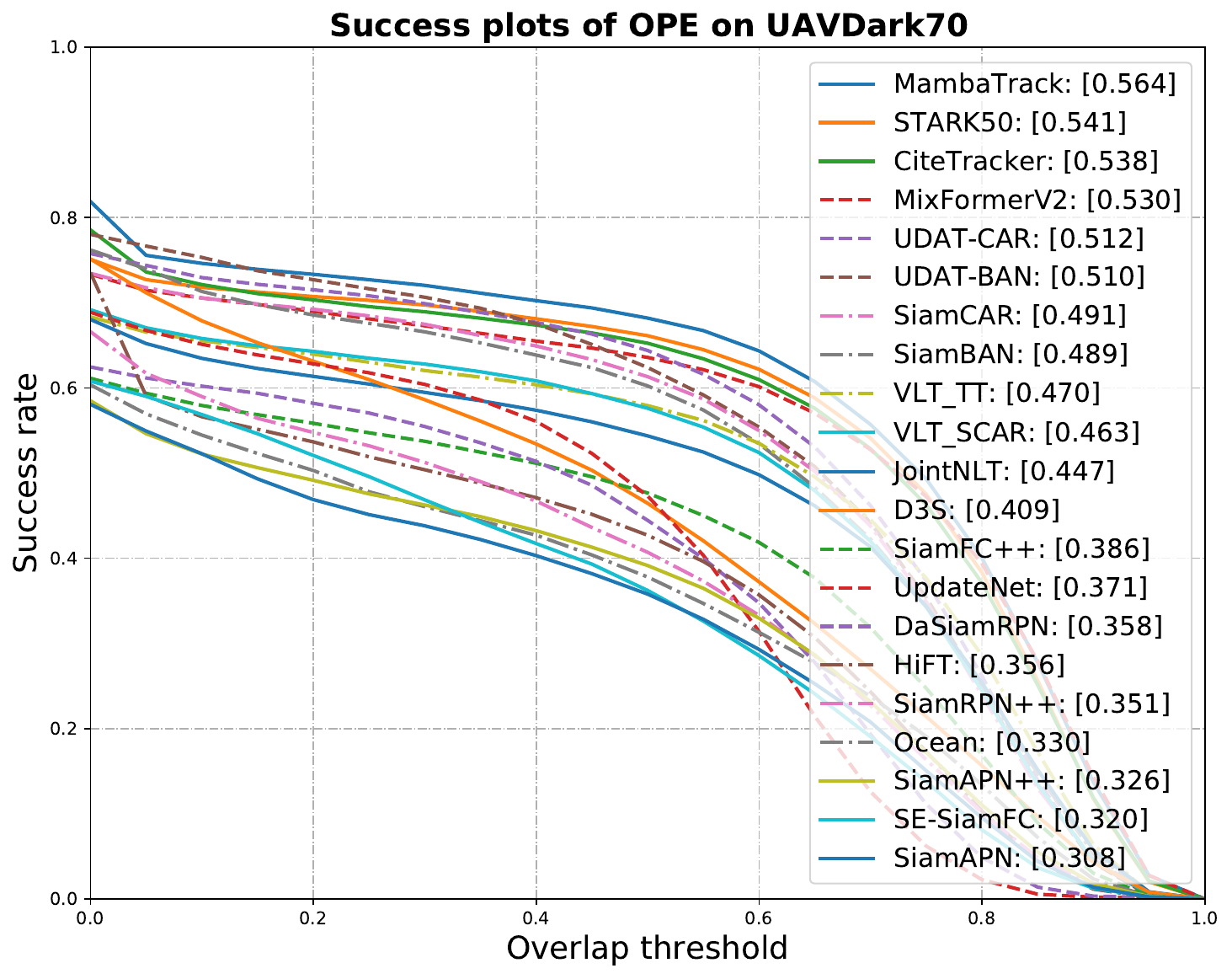}}~
\subfloat{\includegraphics[width =0.5\columnwidth]{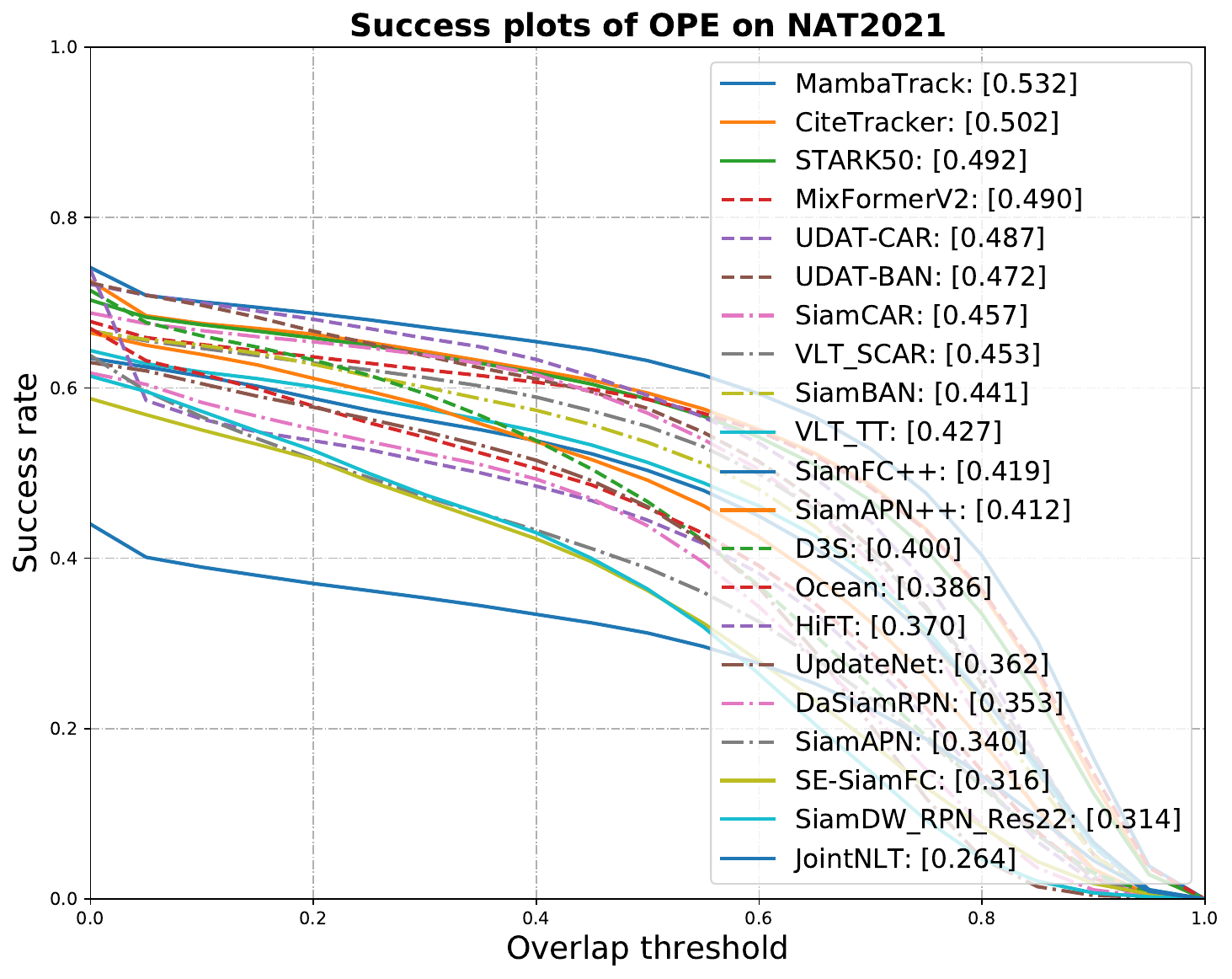}}~
\subfloat{\includegraphics[width =0.5\columnwidth]{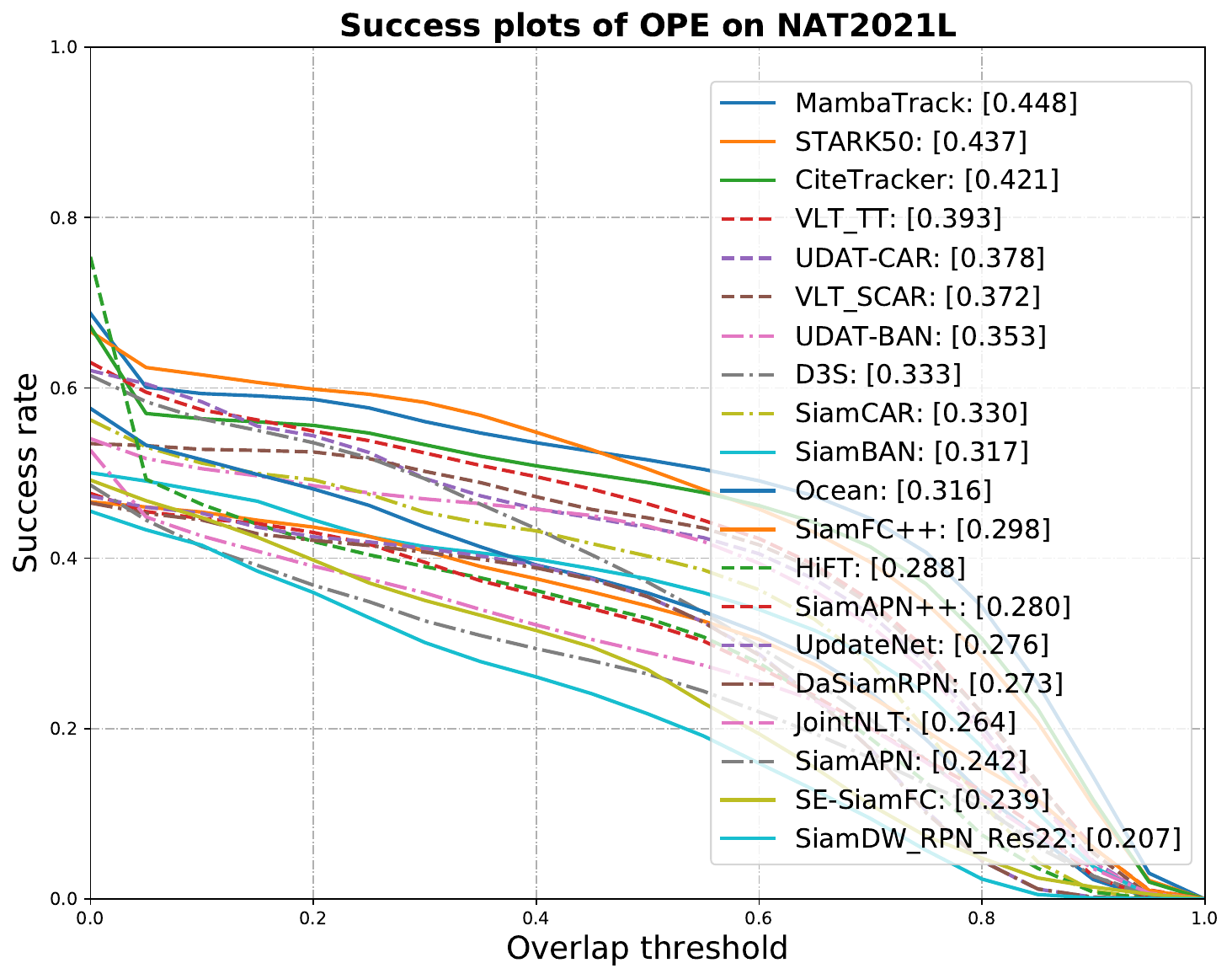}}~
\subfloat{\includegraphics[width =0.5\columnwidth]{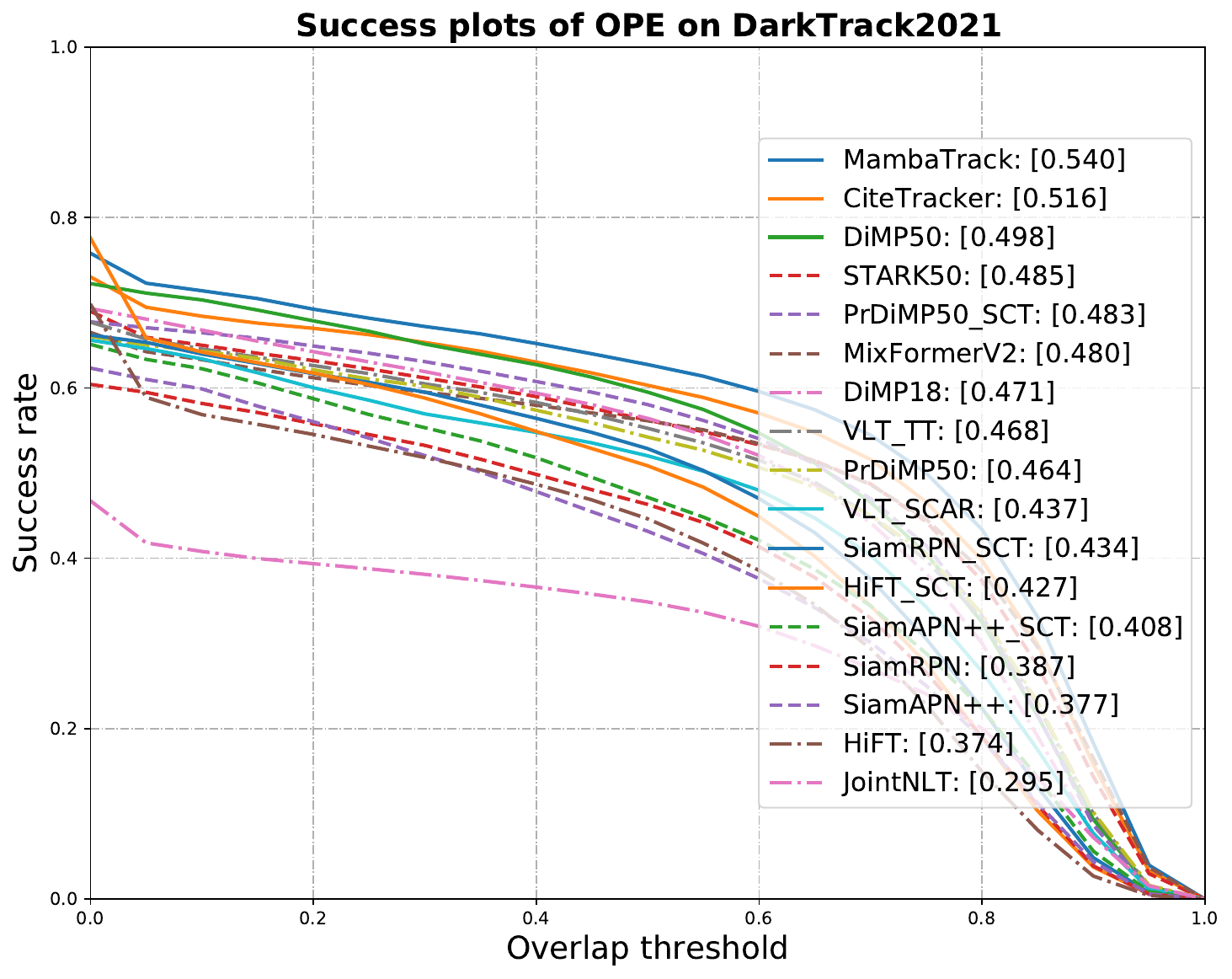}}\\


\caption{Comparison with SOTA trackers on UAVDark70, NAT2021, NAT2021L, and DarkTrack2021 using AUC scores.}
\label{fig:overall_results}
\vspace{-0.4cm}
\end{figure*}

\begin{figure*}[t]
\centering
\centerline{\includegraphics[width=0.85\linewidth]{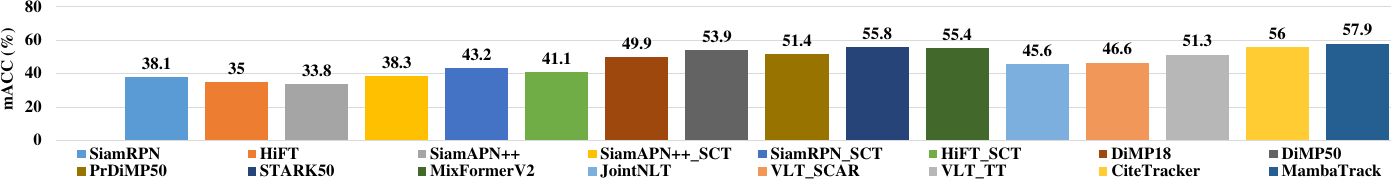}}

\caption{Comparison with SOTA trackers on UAVDark135 using mACC scores. Best viewed in color.}
\label{fig:mACC}
\vspace{-0.5cm}
\end{figure*}

\myPara{Illumination Estimator.} As shown in Eq.~(\ref{eq:mamba-low-light}), we feed the low-light image $I$ and illumination prior $L_{p}$ into the IE to generate a light-up image $I_{lu}$ and its light-up feature map $F_{lu}$. Specifically, the low-light image $I$ and the illumination prior $L_{p}$ are concatenated along the channel dimension, and then fed into three convolutional layers. The first $Conv~1\times 1$ is employed for preliminary feature fusion. The second depthwise separable $Conv~5\times 5$ performs feature upsampling to obtain the light-up feature map $F_{lu}$. Subsequently, we utilize a $Conv~1\times 1$ for downsampling to derive the light-up map $\bar{L}$. Finally, we compute the light-up image $I_{lu}$ as in Eq.~(\ref{eq:simplified}).

\myPara{Damage Restorer.} The DR contains an encoder and a decoder for eliminating noise and artifacts in low-light images~\cite{cai2023retinexformer}. The light-up image $I_{lu}$ is downsampled using a $Conv~3\times3$ to match $F_{lu}$’s dimensions. Two stages of downsampling follow, each combining an illumination fusion state space model (IFSSM)~\cite{bai2024retinexmamba} and a $Conv~4\times 4$ layer. Similar to downsampling, upsampling contains $Deconv~2\times2$, $Conv~1\times1$, and IFSSM layers. Deconvolution doubles the image size and halves the feature dimension, and outputs are concatenated with the corresponding stage of downsampling to \emph{mitigate information loss}. A $Conv~3\times3$ layer restores the image $I_{re}=DR(I_{lu}, F_{lu})$, followed by element-wise addition of $I_{lu}$ to produce the final enhanced image $I_{en}$, as in Eq.~(\ref{eq:mamba-low-light}). \emph{Different from existing works~\cite{cai2023retinexformer,li2021adtrack,zhang2019robust}, our lightweight mamba-based low-light enhancer is used to process the cropped small template and search region, thus achieving efficient image enhancement with low latency.}

\subsection{Cross-modal Mamba Network}

Inspired by~\cite{he2024pan,huang2024mamba}, we introduce a CMM network (see Fig.~\ref{fig:mambatrack}) for vision-language enhancement learning. The core idea is to map visual and language embeddings into a shared space, utilizing a gating mechanism to enhance complementary feature learning and mitigate redundant information, thus optimizing cross-modal feature integration. We first inject language information into the search embeddings through element-wise multiplication, as $\bar{H}_x=H_x\otimes H_t$. Then, we concatenate the template and search embeddings to derive $H_{vl}=[\bar{H}_x;H_z]$. Similarly, we can obtain the corresponding embeddings $H_{v}$ without language injection. Subsequently, we obtain $h_{v}$ and $h_{vl}$ through a normalized layer and a linear projection as follows:
\begin{equation}
h_{m}=Linear_{m}(Norm_{m}(H_{m})),~~~m\in \{v, vl\}.
\label{eq:normlized-embeddings}
\end{equation}
Subsequently, we compute $y_{v}$ and $y_{vl}$ via a 1D convolution~\cite{kiranyaz20211d}, a SiLU activation function~\cite{elfwing2018sigmoid}, and a state space model (SSM)~\cite{gu2023mamba} as follows: 
\begin{equation}
y_{m}=SSM(SiLU(Conv(f_{m}))).
\label{eq:activated-embeddings}
\end{equation}
Next, we calculate the refined features $z_{v}$ and $z_{vl}$ through the gating mechanism~\cite{dauphin2017language} as follows:
\begin{equation}
z_{m}=y_{m}\otimes SiLU(f_{v}).
\label{eq:refined-embeddings}
\end{equation}
Finally, $z_{v}$ and $z_{vl}$ are added element-wise, followed by a linear projection, and then residual on $H_{v}$ to generate the fusion feature $\widetilde{H}_{vl}=[\widetilde{H}_x;\widetilde{H}_z]$, where $\widetilde{H}_x$ and $\widetilde{H}_z$ denote language-enhanced search and template embeddings.

\subsection{Tracking Head and Loss}

We borrow the tracking head from~\cite{ye2022joint}, containing a classification head and a bounding box regression head. To speed up model convergence, we adopt multi-task losses, including $\mathcal{L}_1$ loss, GIoU loss $\mathcal{L}_{GIoU}$~\cite{rezatofighi2019generalized} and focal loss $\mathcal{L}_{focal}$~\cite{lin2017focal}, and the total loss can be written as:
\begin{equation}
\mathcal{L} = \lambda_{1}\mathcal{L}_1+\lambda_{GIoU}\mathcal{L}_{GIoU}+\lambda_{focal}\mathcal{L}_{focal},
\label{eq:losses}
\end{equation}
where $\lambda_{1}$, $\lambda_{GIoU}$, and $\lambda_{focal}$ are balance factors.

\section{Experiments}
\label{sec:experiments}
\subsection{Implementation Details}

Our experimental platform is a server with 8 RTX A6000 GPUs. We adopt Vim-S~\cite{zhu2024vision}, GPT-NeoX~\cite{gpt-neox-library}, and Mamba-130M~\cite{gu2023mamba} as visual encoder, text tokenizer and language encoder, respectively. The template and search region are  $2^{2}$ and $4^{2}$ times of the target and then resized to $128\times 128$ and $256\times 256$. The balance factors $\lambda_{1}$, $\lambda_{GIoU}$, and $\lambda_{focal}$ are set to 5, 2, and 1.5. The visual and language dimensions are 384 and 768. The low-light enhancer is trained using LOL~\cite{wei2018deep}. The tracker is trained using the training splits of TrackingNet~\cite{muller2018trackingnet}, GOT-10k~\cite{huang2019got}, LaSOT~\cite{fan2019lasot}, COCO~\cite{lin2014microsoft}, and WebUAV-3M~\cite{webuav3m2023tpami}. We evaluate different trackers on five night UAV tracking datasets, \ie, DarkTrack2021~\cite{ye2022tracker}, NAT2021~\cite{ye2022unsupervised}, NAT2021L~\cite{ye2022unsupervised}, UAVDark70~\cite{li2021adtrack}, and UAVDark135~\cite{li2022all}. For fair comparisons, we primarily adopt tracking results of methods from original datasets~\cite{ye2022tracker,ye2022unsupervised,li2021adtrack,li2022all}. Due to the absence of language annotations in these night UAV tracking datasets, we manually annotate 518 language prompts (each video with a sentence describing the class name of the target, its attributes, and the surrounding environment) for multimodal night UAV tracking. \emph{\textbf{To facilitate future research, we will release these language annotations.}} 

\vspace{-0.1cm}
\subsection{Comparison with SOTA Methods}
\vspace{-0.1cm}
We evaluate MambaTrack on five challenging night UAV tracking benchmarks, \ie, NAT2021, NAT2021L, DarkTrack2021, UAVDark70, and UAVDark135, which contain 180, 23, 110, 70, and 135 videos, respectively. Among them, NAT2021L is a long-term tracking dataset. Alongside algorithms reported in original datasets, we evaluate multiple current SOTA trackers, \eg, STARK50~\cite{yan2021learning}, MixFormerV2~\cite{cui2024mixformerv2}, JointNLT~\cite{zhou2023joint}, VLT$\rm_{TT}$~\cite{guo2022divert}, and CiteTracker~\cite{li2023citetracker}. Figs.~\ref{fig:overall_results} and~\ref{fig:mACC} indicate that MambaTrack achieves the best AUC and mACC scores among all compared trackers.

\begin{table}[t]

    \setlength{\abovecaptionskip}{0pt}
    \setlength{\belowcaptionskip}{0pt}
	\caption{Ablation study of our approach on NAT2021 and UAVDark135 datasets.}
\vspace{-0.2cm}
    \label{tab:ablation_study}
   {
	\setlength{\tabcolsep}{3.7pt}
	\begin{center}
	\begin{tabular}{l|ccc|ccc}

        \hline
        \multicolumn{1}{l|}{\multirow{2}[1]{*}{Method}} & \multicolumn{3}{c|}{{NAT2021}} & \multicolumn{3}{c}{UAVDark135} \\
			 
        \cline{2-4} \cline{5-7} \multicolumn{1}{c|}{}  & AUC &  $P$  & $P_{ norm}$ & AUC & $P$  & $P_{ norm}$ \\
 
	\hline
  
         Baseline & 48.7 & 63.7 & 51.3  & 51.4  & 61.1  & 54.5 \\
            
         Baseline + MLLE & 51.4 & 66.0 & 54.3  & 54.5  & 65.1  & 58.5 \\
                        
         Baseline + MLLE + CMM  & \textbf{53.2} &   \textbf{67.5} & \textbf{56.8} &  \textbf{57.2} &  \textbf{67.6}  &  \textbf{61.4}
         \\
	   \hline

	\end{tabular}
	\end{center}
 }
\vspace{-0.4cm}
\end{table}

\begin{table}[t]

    \setlength{\abovecaptionskip}{0pt}
    \setlength{\belowcaptionskip}{0pt}
       \caption{Attribute-based evaluation of MambaTrack and other SOTA visual-based and VL-based methods on UAVDark135 dataset using AUC scores.}
   \vspace{-0.2cm}
	\label{tab:attribute_evaluation}
	\setlength{\tabcolsep}{5.1pt}
	\begin{center}
	\begin{tabular}{l|c|ccccc}
	
	\hline
	Method &  Type	&  FM   & IV  & LR  & OCC &  VC\\
        \hline  
        HiFT~\cite{cao2021hift} &  Visual-based &  24.1 &   19.1 & 14.7 &  13.0 &  15.1 \\
        SiamRPN~\cite{li2018high} &  Visual-based &  27.1 &   19.6 &  14.9 &  16.1  &  15.5 \\
        DiMP50~\cite{bhat2019learning} &  Visual-based &  38.0 &   29.8 &  21.6 &  20.4  &  23.9 \\
        MixFormerV2~\cite{cui2024mixformerv2} &  Visual-based &  39.4 &   \textbf{31.2} &  \textbf{22.3} &  \textbf{21.4}  &  \textbf{24.3} \\
        STARK50~\cite{yan2021learning} &  Visual-based &  \textbf{39.9} &   30.5 &  22.0 &  21.3  &  24.0 \\   
        \hline
    
        JointNLT~\cite{zhou2023joint} &  VL-based &  33.3 &  26.6 &  19.5 &  16.9  &  19.3 \\
        VLT$_{\rm SCAR}$~\cite{guo2022divert} &  VL-based &  32.6 &   25.6 &  18.3 &  17.1  &  19.9 \\
        VLT$_{\rm TT}$~\cite{guo2022divert} &  VL-based &  36.8 &   27.5 &  19.4 &  20.3  &  23.2 \\
        CiteTracker~\cite{li2023citetracker} &  VL-based &  40.4 &   30.8 &  22.5 &  21.5  &  24.7 \\
        
        MambaTrack (Ours) &  VL-based &  \textbf{41.5} &   \textbf{32.2} &  \textbf{23.2} &  \textbf{21.9}  &  \textbf{25.4} \\  

		\hline
		\end{tabular}
	\end{center}
\vspace{-0.8cm}
\end{table}

\begin{figure}[t]
\vspace{-0.05cm}
\centering
\centerline{\includegraphics[width=1.0\linewidth]{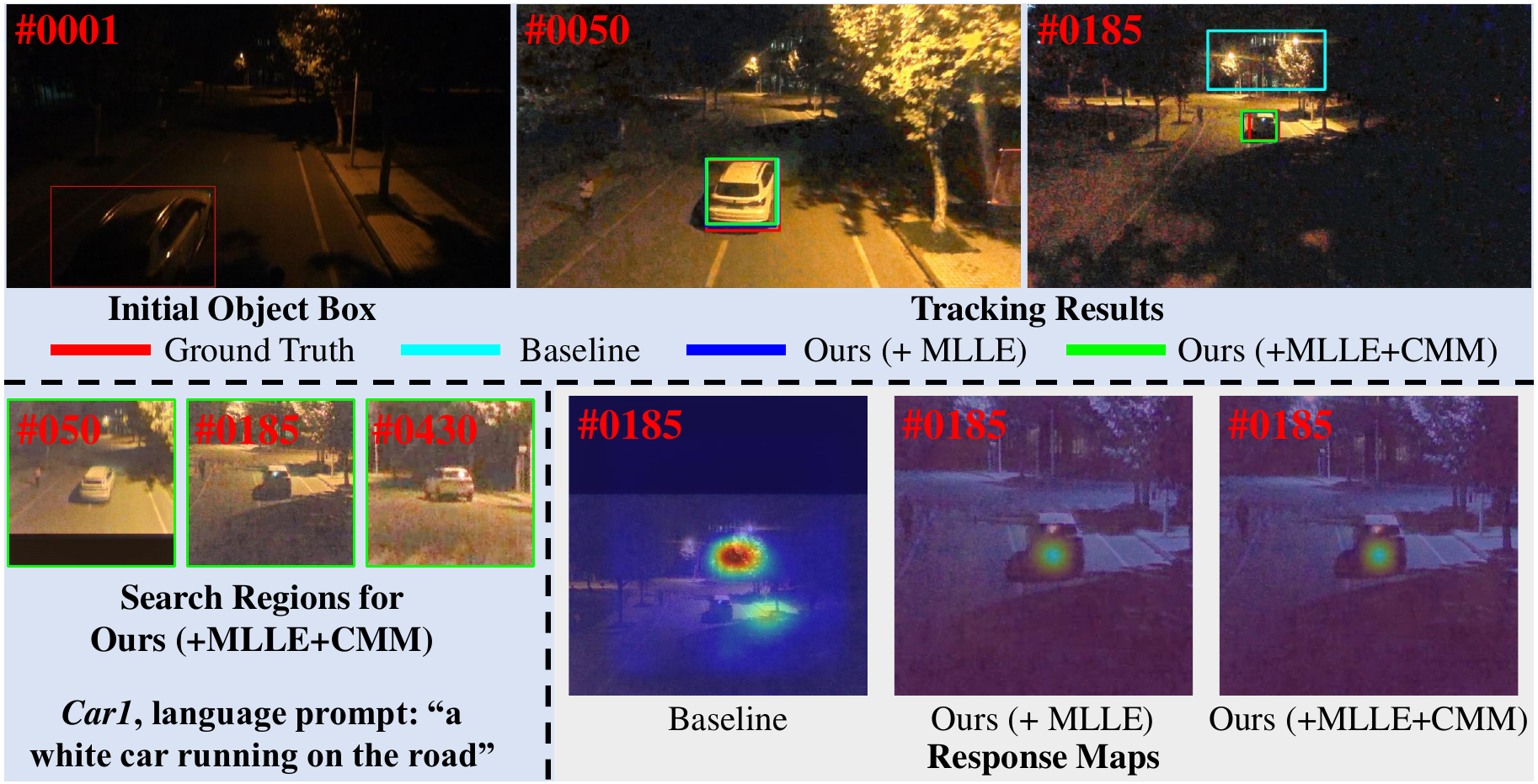}}

\caption{ Visualization of the proposed two components (\ie, MLLE and CMM). The images are enhanced for visualization except for the initial frame. Best viewed by zooming in. }
\label{fig:visualization}
\vspace{-0.2cm}
\end{figure}

\begin{table}[t]
    \setlength{\abovecaptionskip}{0pt}
    \setlength{\belowcaptionskip}{0pt}
       \caption{Comparison of performance, learnable/total parameters, inference speed, and GPU memory with SOTA methods on UAVDark135 dataset using a single RTX A6000.}
 \vspace{-0.2cm}
	\label{tab:model_size_and_speed}
	\setlength{\tabcolsep}{2.5pt}
	\begin{center}
	\begin{tabular}{l|cccc}
	
	\hline
	Method &  AUC	 &  Parameters (M) &   Speed (FPS) &  Memory (MB) \\
        \hline  

        STARK50~\cite{yan2021learning} &  55.0 &   28.2/28.2 &  40  &  1736  \\
        \hline
        
        JointNLT~\cite{zhou2023joint} & 45.1 &   124.3/153.0 &  28  &  3892 \\
        
        VLT$_{\rm TT}$~\cite{guo2022divert} &  50.6 &  100.9/\textbf{100.9} &  30  &  2746 \\
        CiteTracker~\cite{li2023citetracker} &  55.3 &   176.3/176.3 &  11  &  2295 \\
        MambaTrack (Ours) & \textbf{57.2} &   \textbf{15.9}/150.5 &  \textbf{42}  &  \textbf{1142} \\
        
		\hline
		\end{tabular}
	\end{center}
 \vspace{-0.7cm}
\end{table}

 \vspace{-0.2cm}
\subsection{ Ablation Study}
\vspace{-0.1cm}
\myPara{Component-wise Analysis.} To explore the impact of the two crucial components, (\ie, MLLE and CMM) in our tracker, we conduct ablation studies on NAT2021~\cite{ye2022unsupervised} and UAVDark135~\cite{li2022all}. Our baseline includes only the visual branch and the tracking head. We train it with the same settings as MambaTrack to ensure a fair comparison. As shown in Tab.~\ref{tab:ablation_study}, our baseline obtains 48.7\%/63.7\%/51.3\% on NAT2021 and 51.4\%/61.1\%/54.5\% on UAVDark135 in terms of AUC/$P$/$P_{ norm}$ scores. The MLLE module significantly improves the robustness of tracking in the dark, with AUC scores increasing by 2.7\% and 3.1\% on two benchmarks. The CMM module further improves the tracking accuracy by adding semantic information, and the AUC scores increase by 1.8\% and 2.7\% respectively.

\myPara{Attribute-based Performance.} To comprehensively analyze the performance of MambaTrack across various complex scenarios, we report attribute-based evaluation results in Tab.~\ref{tab:attribute_evaluation}. Specifically, we use all five attributes from UAVDark135, which are fast motion (FM), illumination variation (IV), low resolution (LR), occlusion (OCC), and viewpoint change (VC). The results demonstrate that MambaTrack outperforms all compared visual-based and vision-language (VL)-based trackers on the five challenging scenarios.

\myPara{Visualization.} Fig.~\ref{fig:visualization} presents the initial object box, search regions, tracking results, and response maps of the video \emph{Car1} from UAVDark135. The search regions indicate that MambaTrack can concentrate on the real target in complex nighttime scenarios, \eg, illumination variation and fast motion. This is mainly attributed to the proposed mamba-based low-light enhancer, which achieves global image enhancement while preserving important local details that are beneficial for precise target localization. Additionally, we can observe that MambaTrack, by combining low-light enhancement and language enhancement, achieves more focused responses and more accurate tracking results than baseline under poor lighting conditions and highlights the target area.

\subsection{Speed and Model Size Analysis}

Inference speed and model parameters are two key considerations since UAV platforms usually require low latency and low power consumption~\cite{webuav3m2023tpami,10095787}. As shown in Tab.~\ref{tab:model_size_and_speed}, MambaTrack achieves a real-time inference speed of 42 frames per second (FPS). Compared with current SOTA CNN-Transformer-based trackers (\ie, STARK50~\cite{yan2021learning} and VLT$_{\rm TT}$~\cite{guo2022divert}), and Transformer-based trackers (\ie, JointNLT~\cite{zhou2023joint} and CiteTracker~\cite{li2023citetracker}), our MambaTrack reduces learnable parameters by 43.6\% to 91.0\%. Compared with the current SOTA VL-based method CiteTracker, our GPU memory usage is reduced by 50.2\%.

\section{Conclusion}
\label{sec:conclusion}

This work introduces MambaTrack, a novel tracker leveraging dual enhancement techniques for night UAV tracking. MambaTrack utilizes a mamba-based low-light enhancer and a cross-modal mamba network to improve tracking performance in low-light conditions. Another significant contribution of this work is that we annotate language prompts for existing datasets, constructing a new vision-language night UAV tracking task. Extensive experiments on five benchmarks demonstrate the superiority and efficiency of the proposed approach compared to current SOTA methods. 

{
\small
\myPara{Acknowledgements.} This work was supported by the National Natural Science Foundation of China (No. 62471420 and 62101351), and the Major Project of Technology Innovation and Application Development of Chongqing (CSTB2023TIAD-STX0015).}


\bibliographystyle{IEEEtran}
\bibliography{refs}
\end{document}